\documentclass{article}

\usepackage{microtype}
\usepackage{graphicx}
\usepackage{subcaption}
\usepackage{booktabs}
\usepackage{hyperref}

\usepackage[preprint]{icml2026}
\date{}  

\usepackage{amsmath}
\usepackage{amssymb}
\usepackage{mathtools}
\usepackage{amsthm}
\usepackage[capitalize,noabbrev]{cleveref}

\theoremstyle{plain}

\theoremstyle{definition}

\theoremstyle{remark}

\usepackage[textsize=tiny]{todonotes}

\icmltitlerunning{RDP-Based Layer Selection for Sparse LoRA Fine-Tuning}

\begin{document}

\twocolumn[
  \icmltitle{RDP LoRA: Geometry-Driven Identification for Parameter-Efficient Adaptation in Large Language Models}

  \begin{icmlauthorlist}
    \icmlauthor{Yusuf Çelebi}{}
    \icmlauthor{Yağız Asker}{}
    \icmlauthor{Özay Ezerceli}{}
    \icmlauthor{Mahmoud ElHussieni}{}
    \icmlauthor{Selva Taş}{}
    \icmlauthor{Reyhan Bayraktar}{}
    \icmlauthor{Fatma Betül Terzioğlu}{}
  \end{icmlauthorlist}

  \icmlcorrespondingauthor{Yusuf Çelebi}{yusuf.celebi@institution.edu}

  \icmlkeywords{Large Language Models, LoRA, Parameter-Efficient Fine-Tuning, Ramer-Douglas-Peucker, Embedding Trajectories, Layer Selection}

  \vskip 0.3in
]

\printAffiliationsAndNotice{}

\begin{abstract}
Fine-tuning Large Language Models (LLMs) remains structurally uncertain despite
parameter-efficient methods such as Low-Rank Adaptation (LoRA), as the layer-specific
roles of internal representations are poorly understood, leading to heuristic
decisions about where adaptation should be applied. We model the evolution of hidden
states as a high-dimensional geometric trajectory and propose using the
Ramer–Douglas–Peucker (RDP) algorithm, a parameter-free and training-free polygon
simplification method that preserves global structural transitions while eliminating
locally redundant changes, to identify critical breakpoints along the representation
path. Crucially, we use these geometric pivots not merely for analysis, but as a
direct decision signal for determining which layers should be adapted during
parameter-efficient fine-tuning. By integrating this geometry-aware layer selection
strategy into LoRA fine-tuning of Qwen3-8B-Base, we achieve superior performance on
MMLU-Math using only 13 RDP-selected layers (81.67\%), significantly outperforming
both full 36-layer adaptation (79.32\%) and random 13-layer selection (75.56\%), as
well as the baseline Qwen3-8B-Base model (74.25\%). These results demonstrate that
leveraging the intrinsic geometry of representation trajectories provides a robust,
interpretable, and training-free signal for optimizing layer selection during model
adaptation.
\end{abstract}

\section{Introduction}
\label{sec:introduction}

Fine-tuning Large Language Models (LLMs) to adapt to
specific domains is often computationally prohibitive.
Although Parameter-Efficient Fine-Tuning (PEFT)
techniques such as Low-Rank Adaptation (LoRA) \cite{hu2022lora}
alleviate this issue by learning
low-rank updates, they typically apply adaptation
uniformly across layers. This uniform treatment
overlooks a fundamental property of deep networks:
different layers play distinct geometric and functional
roles within the model hierarchy \cite{tenney2019bert}.

We introduce a geometry-based, training-free method to layer selection that assumes the sequence of hidden forward pass states is given as a high-dimensional trajectory. By leveraging the Ramer–Douglas–Peucker (RDP) algorithm (\cite{douglas1973algorithms}; \cite{ramer1972iterative}), a classical curve simplification approach from cartography and visual computing, we isolate structural pivot points along this trajectory. Due to the distance-based nature of this formulation, RDP is also dimension-agnostic and works exactly on low-dimensional paths and high-dimensional embedding sequences, enabling it to preserve global structural transitions while suppressing locally redundant variations.

The motivation for using RDP for this context comes from its dependence on distance-based deviation as its sole operational principle. In modern language models, distances in embedding space are recognized, which encode semantic similarity, with larger geometric separations corresponding to broader semantic transformation. This aligns RDP’s concept of geometric deviation to faithfully represent semantic change, making it a principled technique for locating layers associated with meaningful representational shifts. Most importantly, these geometric pivots are not just something we use as a post-hoc analysis method, but a direct decision signal to help decide which layers should be adjusted during parameter-efficient fine-tuning. Layers corresponding to significant structural variations in the hidden-state trajectory are seen as sites of significant representational shift and are thus prioritized in adaptation.

The coupling of this RDP-based structural signal with velocity-aware Reasoning Band analysis isolates the top semantically active layers for targeted adaptation. Experiments on Qwen3-8B-Base show our Geometry-Selected Sparse LoRA can reach 81.67\% accuracy on MMLU-Math. With only 13 selected layers, it significantly outperforms the full 36-layer LoRA adaptation (79.32\%) and random sparse selection (75.56\%), while modifying substantially fewer parameters. These findings show the embedding trajectories have an intrinsic geometry that can be exploited as a robust, interpretable, and training-free signal for optimizing model adaptation.

\section{Related Work} \label{sec:related_work}

We bring advances to Parameter-Efficient Fine-Tuning (PEFT) using the geometric evolution of the hidden states as a direct, training-free decision signal for model adaptation. Although standard methodologies such as LoRA \cite{hu2022lora} and QLoRA \cite{dettmers2023qlora} apply updates uniformly, recent sparse strategies \cite{wu2024lorasp, kopiczko2024vera} or fusion methods \cite{wang2024loraflow} usually use heuristics, randomization, or module integration, without a principled foundation in the model's intrinsic structure. In our effort to bridge the gap between static analysis and dynamic parameter allocation, we identify high-curvature transitions in the layer-wise trajectory, distinguishing our approach as one that is oriented beyond random sparsity toward intrinsic layer importance.

Expanding upon the view that semantic information is geometrically encoded \cite{Valeriani2023, Lee2025}, we take layer-wise representations in the sense of a continuous path such that high-curvature \textbf{turns} indicate a salient semantic shift. To implement this, we implement the Ramer-Douglas-Peucker (RDP) algorithm \cite{ramer1972iterative}, commonly applied to polyline simplification. Here, RDP has two important components: it is a filtering mechanism that removes local redundancies (noise) and keeps the global \textbf{structural skeleton} of computation \cite{SongZhong2023}, which delivers a strong dimension-neutral signal for choosing the best semantically dense layers, and without any training.

\section{Methodology}
\label{sec:methodology}

Our geometry-driven methodology interprets transformer hidden-state sequences as high-dimensional trajectories to identify structurally significant layers for adaptation. We decouple layer selection from parameter optimization by performing initial forward passes on MMLU-Math subsets to collect hidden states without updating parameters. The \textbf{Ramer–Douglas–Peucker (RDP)} algorithm is then applied to these trajectories to extract ``structural pivots''—layers where the model undergoes its most significant semantic transformations during mathematical reasoning.

Following identification, we freeze the non-essential layers and apply sparse \textbf{LoRA-based fine-tuning} exclusively to the selected pivots. Training is conducted using the \textbf{OrcaMath} \cite{mitra2024orca} dataset to enhance reasoning capabilities. This approach ensures that layer selection is governed by the model's intrinsic representational behavior on benchmarks, while parameter learning focuses on high-quality reasoning data.
\subsection{Problem Definition and Setup}

Consider a pretrained transformer model consisting of $L$ layers. For a given input sequence, the model generates a sequence of hidden representations $V = \{v_1, v_2, \dots, v_L\}$, where each $v_l \in \mathbb{R}^D$ denotes the representation at layer $l$ within a $D$-dimensional embedding space. Our primary objective is to identify a sparse subset of layers that capture the most significant representational transitions throughout the network's forward pass.

Unlike task-specific or gradient-based methods, we consider the layer selection task as a structural simplification problem for the representation trajectories. Following this framework, we treat sequence $V$ as a discrete geometric curve whose evolution happens in $\mathbb{R}^D$. This formulation allows one to directly infer layer importance at each internal representation level based on the intrinsic geometric characteristics of the model’s internal representations and makes the selection process completely training-free and model-agnostic.

\subsection{Ramer--Douglas--Peucker (RDP) Algorithm}

In this work, the Ramer–Douglas–Peucker (RDP) algorithm is
used as a topological filter to identify structural pivot
points along the high-dimensional trajectory ($\mathcal{T}$),
which represents the inter-layer information flow in the
model. By preserving geometrically salient turns that
characterize the trajectory and discarding low-variance
segments, RDP isolates layers corresponding to meaningful
transformations. To put it simply, a layer with the highest
Euclidean deviation from the reference line that connects
the trajectory endpoints is determined; when this deviation
exceeds a threshold $\epsilon$, it is marked as a structural
pivot and the procedure is applied recursively. The entire
algorithmic procedure is elaborated in
Algorithm~\ref{alg:rdp}. This formulation provides the
conceptual foundation for identifying structurally salient
layers based on geometric deformation along the information
trajectory.

In conventional applications of RDP, the distance threshold
$\epsilon$ is manually specified to control the degree of
trajectory simplification. In this work, rather than relying
on a fixed threshold, we introduce a target-driven variant of
RDP tailored to layer-wise representation trajectories.
The specific mechanism by which $\epsilon$ is determined
automatically, as well as its role in multi-scale structural
analysis, is described in the \emph{3.6.2} section.

\begin{algorithm}[tb]
\caption{Ramer--Douglas--Peucker (RDP)}
\label{alg:rdp}
\begin{algorithmic}[]
\STATE \textbf{Input:} Ordered list of points $P = \{p_1, \dots, p_n\}$, distance threshold $\epsilon$
\STATE \textbf{Output:} Simplified list of points $P'$
\STATE $d_{max} \leftarrow 0$
\STATE $index \leftarrow 0$
\STATE $n \leftarrow \text{length}(P)$
\FOR{$i = 2$ \textbf{to} $n-1$}
    \STATE $d \leftarrow \text{perpendicularDistance}(p_i, \text{Line}(p_1, p_n))$
    \IF{$d > d_{max}$}
        \STATE $index \leftarrow i$
        \STATE $d_{max} \leftarrow d$
    \ENDIF
\ENDFOR
\IF{$d_{max} > \epsilon$}
    \STATE $res1 \leftarrow \text{RDP}(P[1 \dots index], \epsilon)$
    \STATE $res2 \leftarrow \text{RDP}(P[index \dots n], \epsilon)$
    \STATE \textbf{return} $\text{concatenate}(res1[1 \dots \text{end}-1], res2[1 \dots \text{end}])$
\ELSE
    \STATE \textbf{return} $\{p_1, p_n\}$
\ENDIF
\end{algorithmic}
\end{algorithm}

\begin{figure}[h] 
\centering 
\includegraphics[width=1\linewidth]{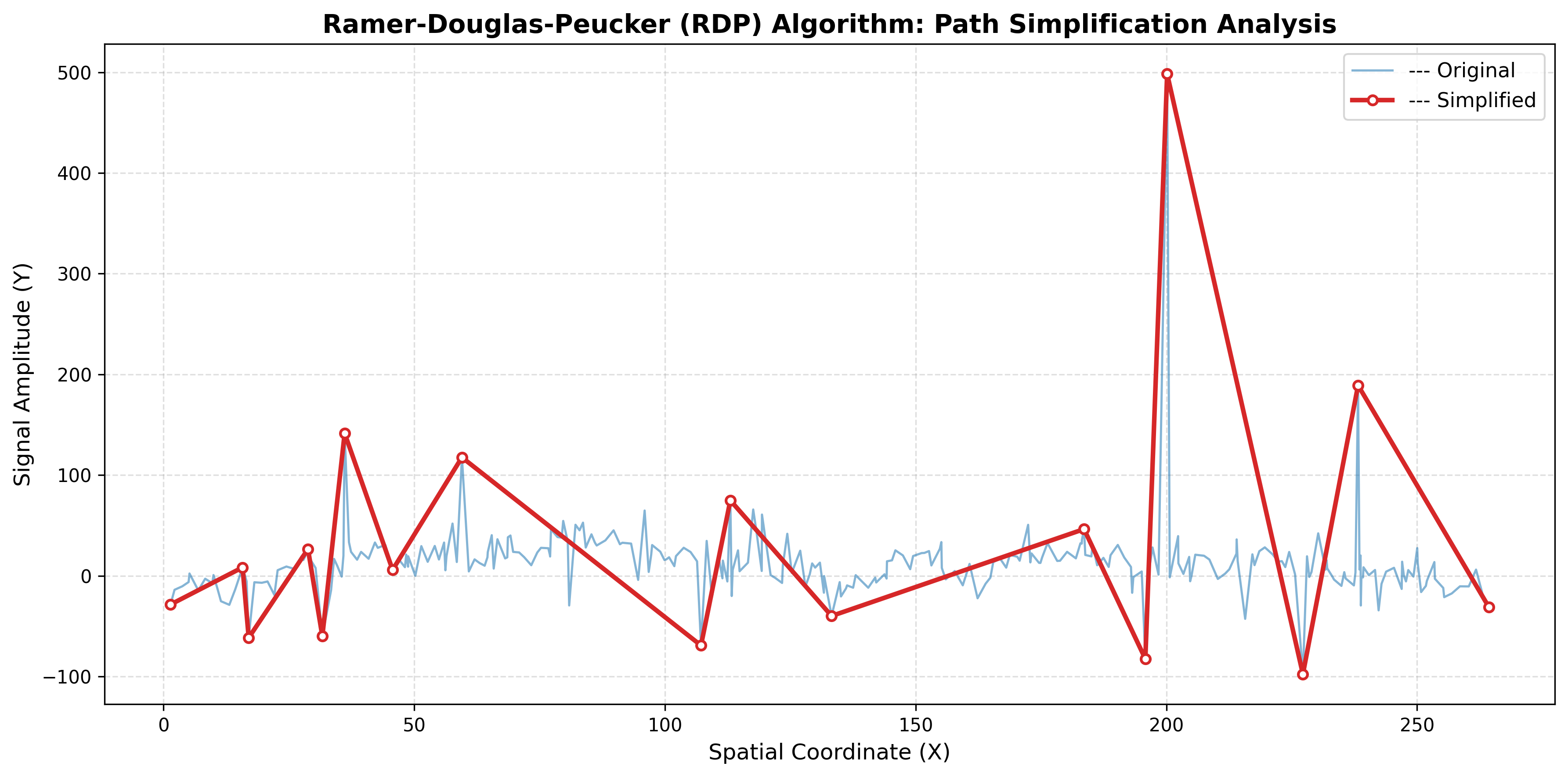} 
\vspace{-5pt}
\caption{Ramer–Douglas–Peucker (RDP) Algorithm: Noise suppression and identification of structural pivot points in a 2D signal.} 
\label{fig:rdp-2d} 
\vspace{-10pt}
\end{figure}

By applying RDP, increasing the threshold value results in progressively coarser approximations of the curve, suggesting that the extent to which the method imposes structural abstraction is clearly modifiable. The algorithm’s operation on a 2D signal, as demonstrated in Figure \ref{fig:rdp-2d}, shows that RDP can suppress noise-like micro-oscillations while maintaining the dominant geometric structure, exposing the underlying structural backbone of the trajectory.

\subsubsection{Dimension-Agnostic Structure and Extension to High-Dimensional Data}
\label{sec:dim_agnostic}

\begin{figure}[tb]
    \centering
    \includegraphics[width=1\linewidth]{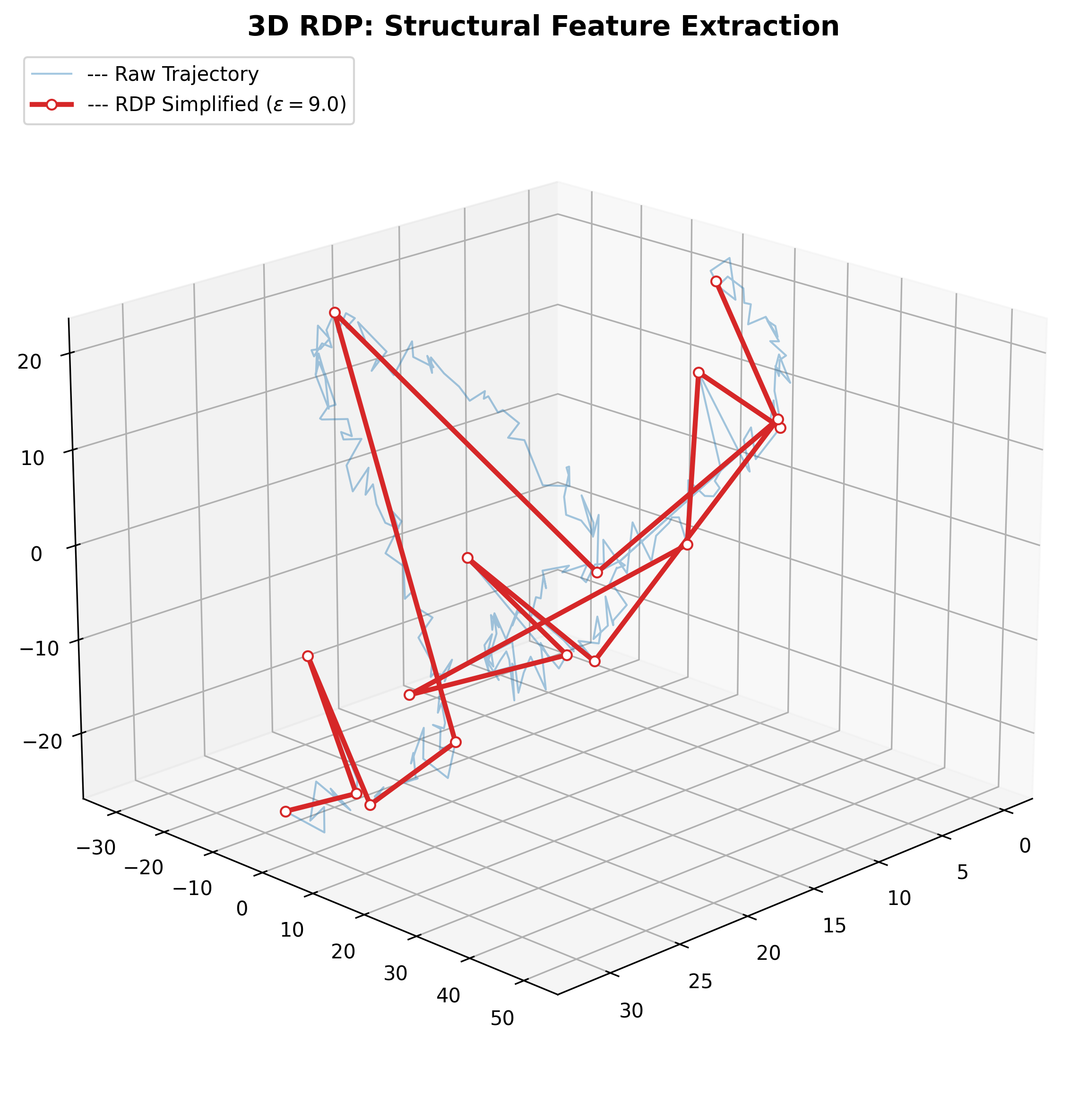}
    \vspace{-5pt}
    \caption{\textbf{Dimension-Agnostic Simplification in 3D.} Visualization of the RDP algorithm applied to a 3D trajectory. The algorithm identifies structural pivots (emphasized points) based on maximum orthogonal deviation, preserving the global topology while filtering local noise.}
    \label{fig:rdp-3d}
\vspace{-25pt} 
\end{figure}

The Ramer-Douglas-Peucker (RDP) algorithm is inherently dimension-agnostic, as its core operation consists solely of computing the orthogonal distance between a point and a reference line. This distance is well-defined in any Euclidean space, allowing RDP to operate identically in 2D, 3D, and high-dimensional embedding spaces without modification (see Figure~\ref{fig:rdp-3d}). Consequently, embedding sequences in modern language models often residing in 768 or higher dimensions can be treated as high-dimensional trajectories to which RDP can be directly applied in a training-free manner. This property enables a seamless transition from low-dimensional geometric intuition to the analysis of semantic structures in embedding spaces, which we examine in the subsequent section.

\subsection{Geometric Interpretation of Embedding Sequences}

Modern language models map each token into a high-dimensional vector space, where training objectives encourage semantically related concepts to be positioned in close proximity \cite{Grand2022,Lee2025}. In other words, semantic similarity is explicitly encoded as geometric distance within the embedding space. This spatial organization of representations constitutes a geometric manifestation of how the model internalizes and structures relationships among concepts \cite{SongZhong2023,Valeriani2023}.

\begin{figure}[ht] 
\centering 
\includegraphics[width=1\linewidth]{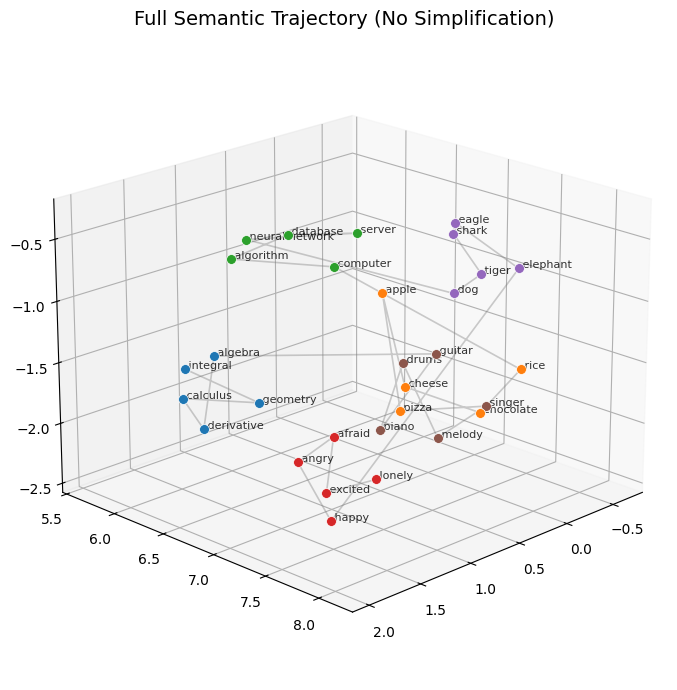} 
\vspace{-5pt} 
\caption{Full Semantic Trajectory: Raw spatial arrangement of distinct conceptual groups (mathematics, music, technology, food, emotions, animals) within the representation space \cite{Valeriani2023,Lee2025}.} 
\label{fig:full-trajectory} 
\vspace{-15pt} 
\end{figure}

In our 3D projection analysis (see Figure~\ref{fig:full-trajectory}), the semantic clustering capacity of the embedding space is clearly observable \cite{SongZhong2023,Lee2025}. Words that refer to different conceptual domains (e.g., mathematics (integral, calculus), animals (\textit{shark, tiger}), music (\textit{melody, guitar})) are clustered into internally coherent clusters that form distinct semantic islands in the representation space. 

This analysis identifies one of the most salient nuances relating to the role of contextual polysemy in geometric positioning within this representation space \cite{Grand2022}. As an instance, although the term \textit{"apple"} denotes a fruit in a literal sense, its embedding stabilizes closer to the tech cluster (i.e., computer, algorithm) than the food-related group. This observation indicates that the prevailing context in the pretraining corpus is significant in determining the geometric coordinates of lexical representations \cite{Grand2022,FreenorAlvarez2025}. 

When embedding sequences are viewed as geometric curves that evolve in high-dimensional space, coherent portions of the resulting trajectory imply semantically coherent regions, and sudden changes of direction or significant shifts signal transitions between separate conceptual areas \cite{SongZhong2023,Valeriani2023}. This sensitivity of geometric position serves as the backbone of the RDP-based trajectory simplification procedure, which is analyzed further in the following section.

\subsubsection{Semantic Skeleton Extraction via RDP and Dimension-Agnosticity}

As previously reported in Section~\ref{sec:dim_agnostic}, the primary function of the RDP algorithm, to compute the orthogonal distance between a point and a reference line, is directly dimension-independent. This property allows the method to generalize its power from a 2D or 3D plane to a modern embedding trajectory that would be found in a 768-dimensional or greater space \cite{Valeriani2023}. Further, since the criterion is used purely based on linear distance, the algorithm predicts information density in a (\textit{training-free}) way without needing any additional parameter learning or optimization.

\begin{figure}[ht]\centering\includegraphics[width=1\linewidth]{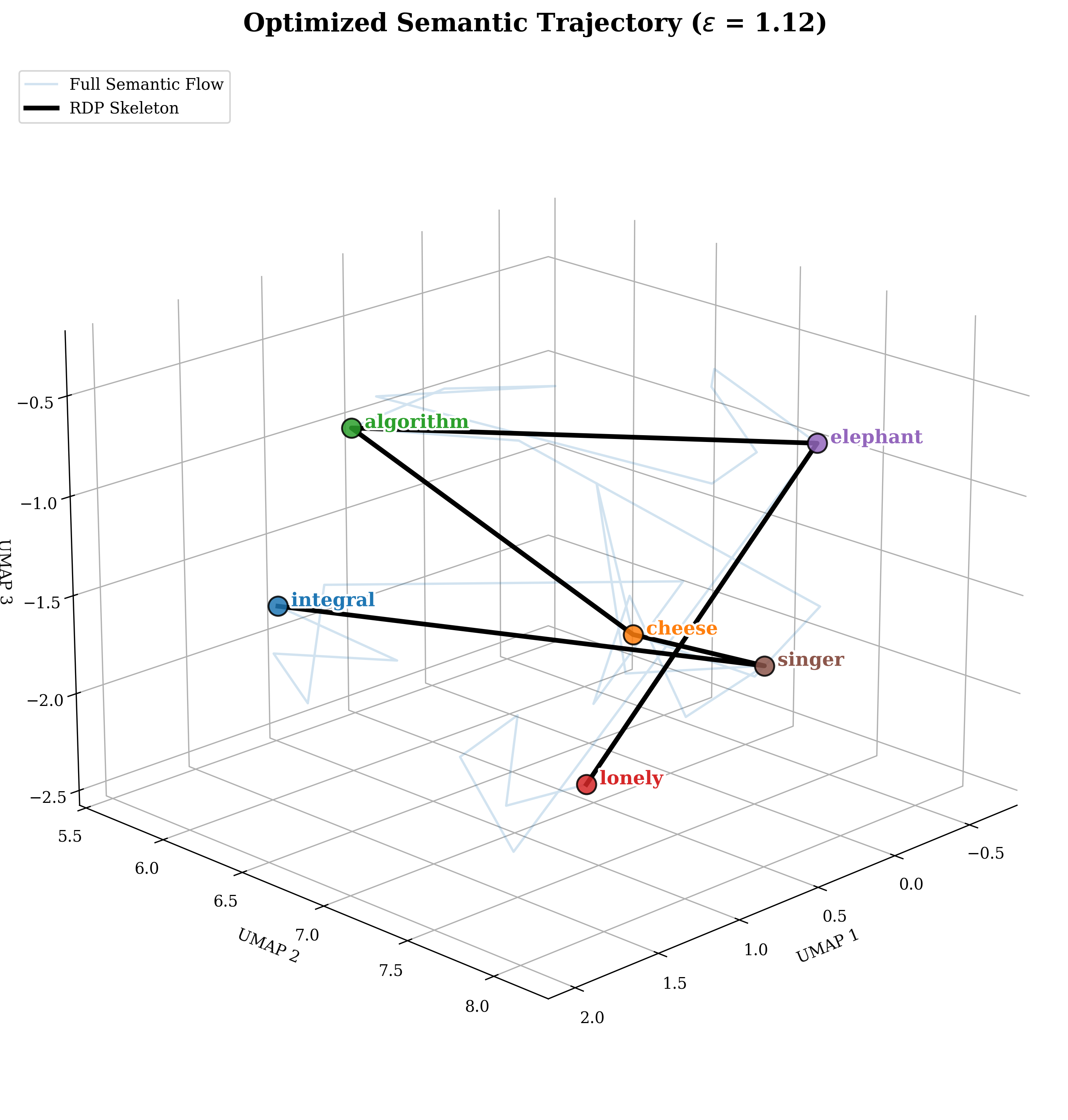}\vspace{-5pt}\caption{Optimized Semantic Trajectory ($\epsilon = 1.12$): By filtering noise in the semantic flow, RDP preserves critical pivot points such as \textit{algorithm, integral, elephant,} and \textit{lonely}, thereby revealing the underlying skeleton of the semantic trajectory.}\label{fig:simplified-skeleton}\vspace{-10pt}\end{figure}

In our experiment, RDP was tested on an embedding trajectory built from a sequence of several semantic categories as mathematics, music, technology, animals, emotions, and food. As illustrated in Figure~\ref{fig:simplified-skeleton}, working with threshold $\epsilon = 1.12$ the algorithm preserves the dominant geometric structure and effectively extracts the principal structural skeleton of the trajectory, whilst removing micro-scale oscillations in the raw trajectory. The results show that RDP’s capacity for retaining structural transformations in high-dimensional space is an effective means to identify semantic transition points, or structural pivots \cite{Lee2025}. Finally, this geometric summarization ability serves as the core of our methodology and the foundation for the next stage, which concerns the analysis of hidden-state trajectories over model layers and how to filter layers for LoRA-based adaptation.

\subsection{Layer-wise Trajectory Extraction and Attention-Weighted Projection}

In the preceding sections, we examined the geometric properties and semantic trajectories of token sequences within a static embedding space. At this stage, we extend the analysis along the model’s depth axis ($L$) by formalizing the dynamic transformation performed across layers for a fixed input $X$ \cite{VanAken2019,VanAken2020}. In a Transformer architecture, each layer $l \in \{1, \dots, L\}$ acts as a non-linear operator that maps its input to a subsequent representation space. However, the layer-wise output matrix $\mathbf{H}_l \in \mathbb{R}^{T \times D}$ is not directly amenable to geometric trajectory analysis; instead, it must be reduced to a single vector $z_l \in \mathbb{R}^D$ that summarizes the semantic state of the layer. In this work, we adopt an Attention-Weighted Projection method to derive layer-level representations. This choice constitutes a deliberate design decision, in contrast to commonly used alternatives in the literature such as mean pooling or last-token representations \cite{VanAken2019}.

While mean pooling strategies risk diluting layer-level signals by aggregating semantically low-information tokens (e.g., stop words and punctuation), relying solely on the hidden state of the final token ($x_T$) fails to capture which contextual elements the model emphasizes at a given depth \cite{VanAken2020}. Owing to the autoregressive nature of causal language models, the token $x_T$ integrates all contextual information processed up to that point. To reflect this contextual refinement within the geometric trajectory, we leverage the attention weights ($\alpha_{l,k}$) that the final token distributes over the entire sequence across $K$ attention heads, using them as an importance-based filtering mechanism \cite{KatzBelinkov2023}.

For each layer $l$, the layer-level representation vector $z_l$ is defined as
\begin{equation}
w_{l,t} = \frac{1}{K} \sum_{k=1}^{K} \alpha_{l,k}(x_T, x_t), \quad
z_l = \sum_{t=1}^{T} w_{l,t} \, h_{l,t}.
\end{equation}

The resulting sequence $\mathcal{T} = \{z_1, z_2, \dots, z_L\}$ constitutes a discrete hidden-state trajectory that evolves along the layer index of the model within a high-dimensional space \cite{VanAken2019,Valeriani2023}. This formulation anchors the notion of a semantic trajectory to the model’s internal operational hierarchy, thereby enabling the RDP algorithm to identify structural transformations occurring across layers as the model processes the input \cite{VanAken2020}.

\subsection{Structural Geometry Analysis and Adaptive Reasoning-Relevant Band Identification}

Information processing across the layers of Transformer architectures does not exhibit a homogeneous distribution. It is well established that early layers primarily map the input into a semantic representation space through feature extraction, whereas later layers are responsible for preparing these representations for the output vocabulary via formatting and logit mapping. The layers situated between these two extremes constitute the core interval in which the model performs complex conceptual transformations and semantic density reaches its peak \cite{Valeriani2023,Lee2025}. We refer to this interval as the \emph{Reasoning-Relevant Band} ($L_{\text{rb}}$).

\begin{figure}[ht]\centering\includegraphics[width=1\linewidth]{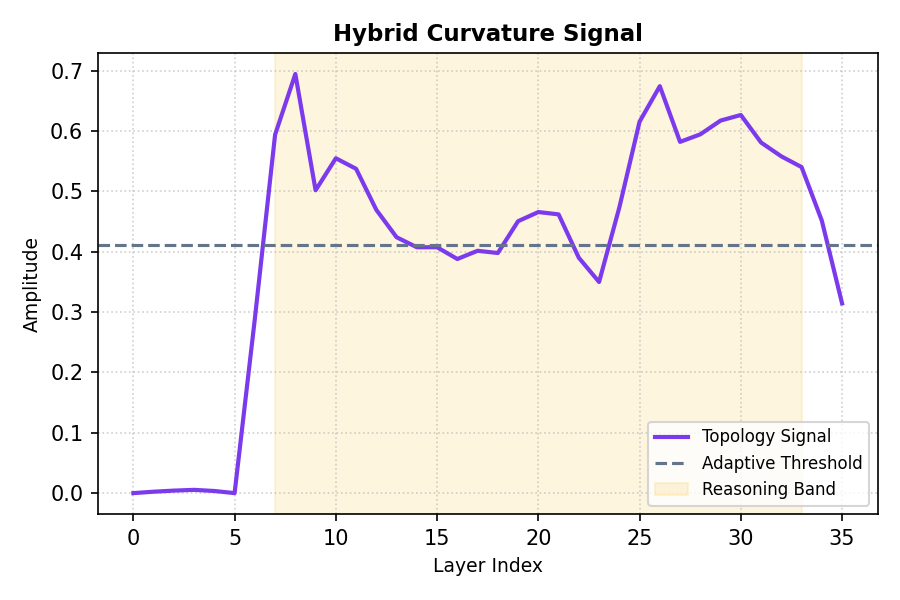}\vspace{-5pt}\caption{\textbf{Topological Hybrid Signal Analysis:} The purple curve represents the computed $S(l)$ signal, the dashed line denotes the adaptive threshold value ($\tau$), and the shaded brown region indicates the identified Reasoning Band interval.}\label{fig:hybrid-signal}\vspace{-10pt}\end{figure}

At this stage of our methodology, we aim to identify this dynamically active interval by defining a hybrid structural signal $S(l)$ over the layer trajectory $\mathcal{T}$, which jointly captures global structure and local dynamics:
\begin{equation}
S(l) = \alpha \cdot \text{Dev}(l) + (1 - \alpha) \cdot \text{Vel}(l),
\end{equation}
where $\text{Dev}(l)$ denotes the Euclidean deviation of the point $z_l$ from the reference line connecting the trajectory endpoints, and $\text{Vel}(l)$ represents the rate of change between successive layer representations. First, the signal is smoothed with a Savitzky--Golay filter to suppress micro-scale oscillations caused by layer-to-layer transitions. Afterwards, an adaptive thresholding mechanism using Otsu’s approach that maximizes inter-class variance is used to identify the band boundaries. The sequence satisfying $S(l) > \tau$ is selected as the semantic core, corresponding to the domain where the model expends the highest degree of structural effort in the representation transformation process \cite{Lee2025} (see Figure~\ref{fig:hybrid-signal}).

The trajectory provided in this research is not a single input instance, but represents the statistical average of Attention-Weighted hidden-state representations computed over all samples of the dataset. This joint approach reduces idiosyncratic input-level divergences, revealing the underlying operational properties of the model architecture \cite{Valeriani2023}. As can be seen from above, the above trajectory shows a generally linear progress in the outer layers along with an increasing topological complexity and semantic clustering in the Reasoning Band. This structural pattern is interpreted as an indication that the model’s reasoning capacity is concentrated within this relatively limited geometric range.

\begin{figure}[h]
\centering
\includegraphics[width=1\linewidth]{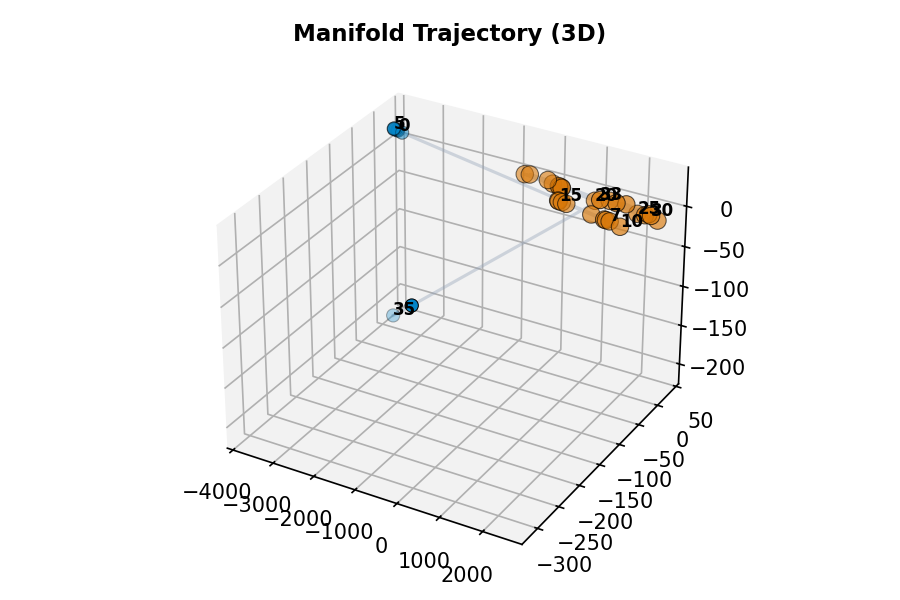}
\vspace{-5pt}
\caption{
\textbf{3D PCA Projection of the Hidden-State Trajectory.}
The trajectory, obtained by averaging across all samples and projected via PCA solely for visualization purposes (without preserving full geometric fidelity), highlights the Reasoning Band (central segment) as the region exhibiting the highest degree of curvature and semantic transformation.
}
\label{fig:pca-trajectory}
\vspace{-10pt}
\end{figure}

\begin{table*}[ht]
\centering
\caption{Model-agnostic layer adaptation strategies.
All strategies share identical training and evaluation settings;
only layer selection and LoRA capacity allocation differ.
Capacity allocation refers to how LoRA capacity is distributed
across the adapted layers (uniform, importance-weighted, or
reduced).}

\label{tab:layer_strategies}
\small
\begin{tabular}{p{4.2cm} p{3.2cm} p{3.6cm} p{3.2cm}}
\toprule
\textbf{Strategy} & \textbf{Adapted Layers} & \textbf{Selection Principle} & \textbf{Capacity Allocation} \\
\midrule
No Adaptation & None & -- & -- \\
Full LoRA & All layers & None & Uniform \\
Geometry Selected Sparse LoRA & Top-$K$ layers & Geometry-driven importance & Uniform \\
Geometry Weighted Sparse LoRA & Top-$K$ layers & Geometry-driven importance & Importance-weighted \\
Reduced Geometry Weighted Sparse LoRA & Top-$K$ layers & Geometry-driven importance & Reduced importance-weighted \\
Inverse Geometry Selection & Complement of Top-$K$ & Geometry-driven (inverse) & Uniform \\
Random Sparse LoRA & Random $K$ layers & Random & Uniform \\
Reasoning-Band LoRA & Band layers & Geometry-derived band & Uniform \\
\bottomrule
\end{tabular}
\vspace{-15pt}
\end{table*}

\subsection{Multi-Scale RDP Analysis: Extracting the Structural Backbone}

Building upon the semantic trajectory simplification framework introduced in the preceding sections, we now apply this principle to the sequence of successive hidden-state trajectories of the model, $\mathcal{T} = \{z_1, \dots, z_L\}$. In this section, we formulate the Ramer--Douglas--Peucker (RDP) algorithm as a multi-scale statistical filtering mechanism operating over a data distribution, with the objective of identifying \emph{structural pivot} layers along the trajectory.

\subsubsection{Distributional Consensus across Domain Samples}

To obtain a noise-reduced and robust characterization of the model’s layer-wise hierarchy, the analysis is conducted not on a single input instance but over a dataset $\mathcal{D}$ that reflects the characteristic properties of the target domain. A separate hidden-state trajectory $\mathcal{T}_i$ is thus extracted for every $i \in \mathcal{D}$ of the sample. This collective method removes stochastic variations induced by individual inputs and reveals the model’s global topological behavior. Then, the Ramer--Douglas--Peucker (RDP) algorithm is used on this ensemble of trajectories, mapping the layer-wise structural density of the model as a collective statistical measure.

\begin{figure}[tb]
    \centering
    \includegraphics[width=1.03\linewidth]{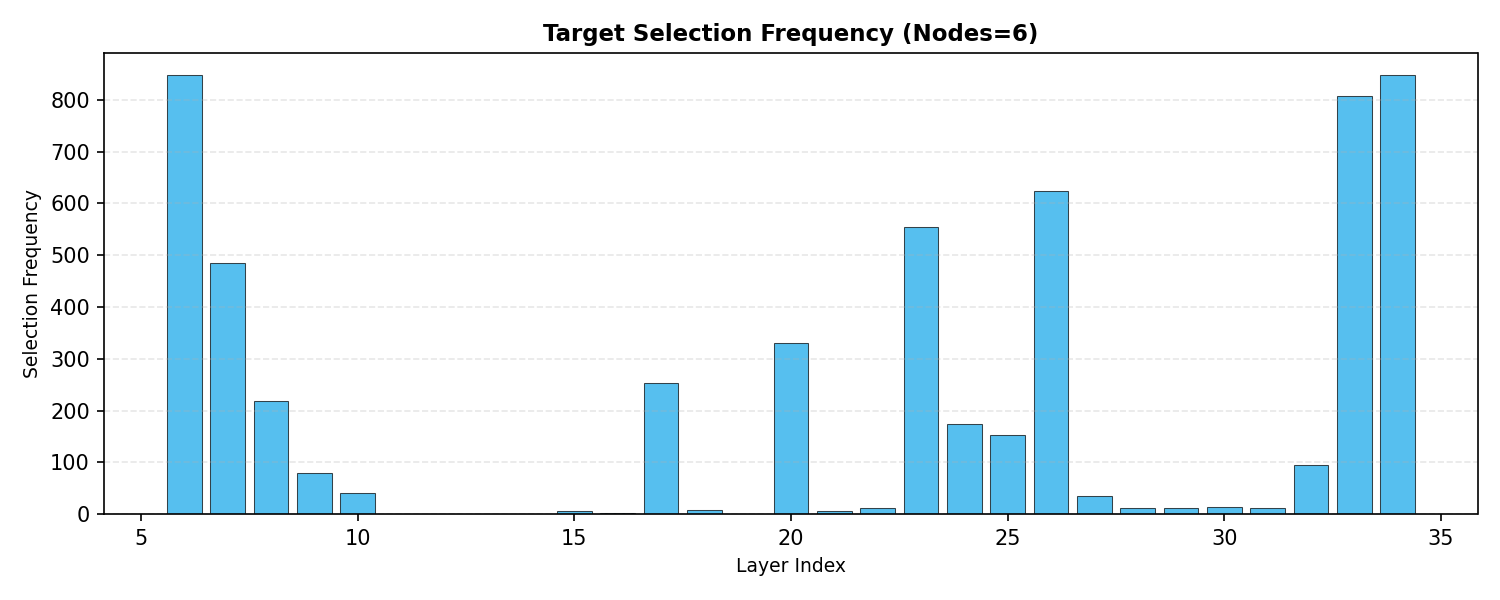}
    \vspace{-5pt}
    \caption{\textbf{Multi-Scale RDP Layer Distribution on Target = 6:} The frequency with which layers are selected as pivots ($t$).}
    \label{fig:rdp-multi-hist}
    \vspace{-25pt}
\end{figure}

Figure~\ref{fig:rdp-multi-hist} demonstrates that RDP consistently targets specific structural subsets, independent of the semantic representation. This stability serves as evidence that the extracted layers capture a universal geometric meaning within the model.

\subsubsection{Multi-Scale Resolution and Dynamic Thresholding}

In a high-dimensional representation space ($D \gg 3$),
a fixed distance threshold ($\epsilon$) is insufficient to
simultaneously capture the global structural skeleton of a
trajectory and its finer local variations. To address this
limitation, we adopt a multi-scale formulation of RDP that
operates over a range of target resolutions rather than a
single fixed threshold.

Importantly, the target resolution $t$ denotes the desired
number of points retained after RDP simplification. Due to
the nature of the RDP algorithm, which always preserves the
first and last points of a trajectory, the minimal meaningful
target is $t=3$, corresponding to the presence of at least
one interior structural pivot. Starting from this minimal
configuration, we progressively increase the target
resolution until the full trajectory is recovered. This
process allows the method to move smoothly from a highly
coarse structural abstraction to increasingly fine-grained
representational detail.

\begin{enumerate}
    \item \textbf{Target-Driven Epsilon Optimization:}
    Rather than manually specifying a distance threshold,
    we invert the conventional RDP formulation and treat
    $\epsilon$ as a dependent variable. For each target
    resolution $t \in T$, the algorithm automatically
    computes the minimal threshold $\epsilon_t$ such that
    the RDP simplification retains at most $t$ points:
    \begin{equation}
        \epsilon_t = \min \{ \epsilon \mid |\text{RDP}(\mathcal{T}, \epsilon)| \leq t \}.
    \end{equation}
    In practice, this is achieved via a monotonic search
    over $\epsilon$, exploiting the fact that the number of
    retained points decreases monotonically as $\epsilon$
    increases. As a result, coarser target resolutions
    (small $t$) correspond to larger $\epsilon_t$, while
    finer resolutions yield progressively smaller thresholds.

    \item \textbf{Multi-Scale Statistical Voting:}
    Structural pivots identified at very small target
    resolutions correspond to dominant global deformations
    of the trajectory and are therefore assigned higher
    importance. However, relying solely on early targets
    risks overlooking layers that contribute consistently
    at finer scales. To balance these effects, we aggregate
    pivot selections across all target resolutions.
    
    Specifically, layers that appear as pivots at early
    targets are naturally emphasized, while layers that
    emerge only at larger targets are not discarded but
    contribute with reduced weight. This yields an
    importance signal that favors globally critical layers
    without suppressing structurally consistent local
    transitions. Formally, we define the accumulated
    \textbf{RDP Importance Score} for each layer $l$ as
    \begin{equation}
        \omega_{RDP}(l) = \sum_{t \in T}
        \frac{\mathbb{I}(l \in \mathcal{P}_t)}{\sqrt{t}},
    \end{equation}
    where $\mathbb{I}(\cdot)$ is the indicator function and
    $\mathcal{P}_t$ denotes the set of pivot layers selected
    at target resolution $t$. The $1/\sqrt{t}$ weighting
    explicitly prioritizes pivots that persist at coarser
    resolutions while retaining sensitivity to finer-scale
    structure, as illustrated in Figure~\ref{fig:rdp-multi-hist}.
\end{enumerate}

\subsection{Geometric Importance Ranking and Adaptive Adaptation Priority}

For our final methodology steps, these multi-size RDP analysis outputs are fed into the Layer Importance Ranking that outlines the information bottlenecks during the forward pass of the model.


We define a \textbf{Structural Importance Index} ($\mathcal{I}_l$) for each layer $l$ in order to incorporate both the global structural skeleton and local dynamics of the trajectory: $\mathcal{I}_l = \beta \cdot \text{norm}(\omega_{RDP}(l)) + (1-\beta) \cdot \text{norm}(Vel(l))$. Here, $\omega_{RDP}(l)$ is the weighted voting score and $Vel(l)$ is the rate of semantic change through layers. The parameter $\beta \in [0, 1]$ dictates the trade-off between critical pivot points at which the trajectory makes substantial shifts in direction and regions of informational acceleration.


The ranking induced by the computed $\mathcal{I}_l$ values
serves as a layer-wise adaptation prior that determines
where and to what extent parameter-efficient fine-tuning
is applied. Rather than enforcing uniform adaptation,
layers with higher geometric importance are prioritized
during training, while less critical layers receive reduced
or no adaptation. The concrete realization of this priority
under different experimental settings is described in
Section~\ref{sec:experiments}.

\section{Experiments}
\label{sec:experiments}

We evaluate our geometric layer selection method (Section~\ref{sec:methodology}) on a high-capacity reference model, \textit{Qwen3-8B-Base}, enabling controlled comparison before assessing transferability to \textit{Qwen3-4B}, \textit{Qwen3-14B}, \textit{DeepSeek-LLM-7B}, and \textit{Gemma-7B}. Primary results are in Table~\ref{tab:qwen8b_math_results}.

\subsection{Layer Adaptation Strategies}
Table~\ref{tab:layer_strategies} summarizes the strategies. All configurations share identical training data and protocols (rank 32, $\alpha=64$ unless noted), differing only in layer selection and capacity allocation.

\subsection{Core Experiments on Qwen3-8B-Base}
We analyze MMLU-Math performance. \textbf{Baselines:} The unadapted model achieves 74.25\%, while uniform \textbf{Full LoRA} reaches 79.32\%.

\textbf{Geometry-Selected Sparse LoRA.} Adapting the $K=13$ pivotal layers (reasoning band 7--33) yields \textbf{81.67\%}, significantly outperforming Full LoRA and validating the geometric selection signal. Here, $K$ reflects a chosen sparsification level rather than a tuned hyperparameter, and is set to approximately half of the identified reasoning-relevant band.

\begin{table}[!t]
\centering
\caption{Math accuracy on \textit{Qwen3-8B-Base} (MMLU-Math).
All strategies share identical training and evaluation settings;
only the layer adaptation strategy differs.}
\label{tab:qwen8b_math_results}
\footnotesize
\begin{tabular}{lc}
\toprule
\textbf{Layer Adaptation Strategy} & \textbf{Math Acc. (\%)} \\
\midrule
No Adaptation (Base) & 74.25 \\
Random Sparse LoRA (Top-$K$) & 75.56 \\
Inverse Geometry Selection & 78.48 \\
Geometry-Weighted Sparse LoRA & 78.20 \\
Reasoning-Band LoRA & 78.10 \\
Full LoRA & 79.32 \\
Reduced Geometry-Weighted Sparse LoRA & 79.23 \\
\textbf{Geometry-Selected Sparse LoRA (Top-$K$)} & \textbf{81.67} \\
\bottomrule
\end{tabular}
\vspace{-15pt}
\end{table}

\textbf{Comparisons.} \textbf{Random Sparse LoRA} ($K=13$) scores only 75.56\%, and \textbf{Reasoning-Band LoRA} (all band layers) reaches 78.10\%, proving that layer identity is more critical than mere sparsity or band constraints. \textbf{Inverse Selection} (non-pivot band layers) yields 78.48\%, further confirming the signal utility. \textbf{Geometry-Weighted} strategies (78.20\%--79.23\%) fell short of uniform allocation on optimal layers, indicating selection matters more than capacity tuning.

All reported results are obtained from a single run; variance analysis across multiple seeds is left for future work.

\subsection{Sensitivity to Model Scale and Architecture}
To assess robustness, we extend our evaluation to \textit{Qwen3-4B}, \textit{Qwen3-14B}, \textit{DeepSeek-LLM-7B}, and \textit{Gemma-7B}. Across all models, geometry-selected sparse LoRA consistently outperforms random baselines. While gains on smaller models (\textit{Qwen3-4B}) are modest, the strategy remains superior to random selection. On larger or distinct architectures (\textit{Qwen3-14B}, \textit{DeepSeek-LLM-7B}, \textit{Gemma-7B}), our method matches or exceeds Full LoRA performance with significantly fewer parameters. Detailed results are provided in Appendix~\ref{app:additional_models}.

\section{Discussion}

\textbf{Layer Selection as a Structural Decision.} Our central finding is that layer selection in LoRA-based adaptation directly determines both performance and efficiency. The representational geometry from the model's forward pass provides a sufficient, training-free signal for identifying critical layers. On Qwen3-8B-Base, performance gains stem not from parameter count but from the structural positions at which adaptation is applied.

\textbf{Layer Identity over Band or Sparsity.} Restricting adaptation to the Reasoning Band alone is insufficient; adapting all band layers or random subsets consistently underperforms RDP-selected pivots. This confirms that adaptation is more sensitive to layer identity than to the number of layers or interval width. Semantic transformations concentrate in a limited subset, making layer selection a qualitative discrimination problem rather than a coverage problem.

\textbf{Selection versus Capacity Tuning}. It is the choice of layers that matters most with respect to capacity allocation. Although asymmetric capacity allocation can be useful in some circumstances, uniform provisioning on geometrically selected layers has relatively robust performance stability. Selection should make up the core design decision and capacity tuning should take a back seat. 

\textbf{Geometry vs. Random Selection.} Even with the same number of adapted layers, geometry-selected layers have consistently better performance than random designs, indicating that RDP captures a relevant structural signal determining adaptation behaviour.

\textbf{Generalization.} Geometry-based selection offers pronounced advantages for medium- and large-scale models. While Full LoRA may achieve marginally higher accuracy on some architectures, our method matches or exceeds it with substantially fewer parameters—an explicit performance-efficiency trade-off.

\textbf{Limitations.} The design space encompassing model families, model scales, and capacity parameters is not exhaustively explored in this study. While layer depth is optimized within the evaluated configurations, the selected capacity settings may not correspond to globally maximal efficiency. Moreover, the empirical analysis is conducted a single benchmark, which may limit the generalizability of the results across architectures and tasks. Consequently, the reported findings should be interpreted as reflecting a limited exploration of the architectural design space rather than indicating a deficiency in the proposed geometric signal. A more systematic and comprehensive investigation of this broader design space is left for future work.

\section{Conclusion}

We presented in this paper a geometry-driven framework for layer selection in LoRA adaptation of Transformer models. The method models hidden-state sequences as high-dimensional representation trajectories and features structurally important layers with the Ramer–Douglas–Peucker (RDP) algorithm. By positioning the choice of layers in the representational geometry generated by the forward pass of the model, the method allows sparse adaptation and avoids dependence on gradients or task-specific training signals. Geometry-selected sparse LoRA  improves over random layer selection and, in most instances, surpasses full adaptation or gets nearly as good performance at substantially fewer trainable parameters. These results suggest that the capability of LoRA-centric fine-tuned approaches for scaling up/down performance is not so much determined by the number of adapted layers as their structural locations in the hierarchy of the given model.

\textbf{Future work.} While this study focuses on a static layer selection regime, the proposed geometric signal naturally suggests extensions to more dynamic adaptation settings. For example, layer importance could be determined in an input-dependent manner during inference, or updated online during training based on batch-level representation geometry. Exploring such dynamic formulations may provide further insight into the relationship between the temporal evolution of representation geometry and parameter efficiency, and help broaden the practical applicability of geometry-driven adaptation.

\newpage

\bibliography{example_paper}
\bibliographystyle{icml2026}

\newpage
\appendix

\section{Appendix: Additional Experimental Results}
\label{app:additional_models}

In this appendix, we provide a unified comparative analysis of the layer adaptation strategies across four additional Large Language Models (LLMs) with varying scales and architectures: \textit{Qwen3-4B-Base}, \textit{Qwen3-14B-Base}, \textit{Gemma-7B}, and \textit{DeepSeek-LLM-7B-Base}.

Table \ref{tab:all_models_math} summarizes the MMLU-Math accuracy for each model under five distinct adaptation settings. These results reinforce the findings observed in the main experiments: geometry-driven layer selection strategies consistently yield competitive performance compared to Full LoRA and outperform random baselines, particularly in larger base models. Notably, \textbf{Geometry-Weighted Sparse LoRA} demonstrates superior adaptability in the \textit{Qwen3-14B} and \textit{DeepSeek-LLM-7B} settings, suggesting that allocating capacity based on geometric importance becomes increasingly effective as model scale grows.

\begin{table}[ht]
\centering
\caption{\textbf{Comparative MMLU-Math Accuracy.} Comparison of layer adaptation strategies across different models. \textit{Top-$K$} selection corresponds to approx. $35\%$ of layers.}
\label{tab:all_models_math}
\vskip 0.15in
\begin{small}
\resizebox{\columnwidth}{!}{
\begin{tabular}{lcccc}
\toprule
\textbf{Strategy} & \textbf{Qwen3-4B} & \textbf{Qwen3-14B} & \textbf{Gemma-7B} & \textbf{DeepSeek-7B} \\
\midrule
No Adaptation & 69.45 & 81.11 & 45.58 & 31.67 \\
Full LoRA & \textbf{70.30} & 81.95 & \textbf{49.62} & 32.05 \\
\midrule
Geometry-Selected & 70.11 & 81.86 & 45.39 & 32.05 \\
Geometry-Weighted & 69.64 & \textbf{82.61} & 48.68 & \textbf{32.99} \\
Random Sparse & 70.02 & 81.50 & 47.09 & 31.86 \\
\bottomrule
\end{tabular}
}
\end{small}
\vskip -0.1in
\end{table}

\end{document}